\documentclass[10pt,twocolumn,letterpaper]{article}

\usepackage{cvpr}              

\usepackage{graphicx}
\usepackage{amsmath}
\usepackage{amssymb}
\usepackage{booktabs}

\usepackage{bm}
\usepackage{multirow}
\usepackage{booktabs}

\usepackage{xr}
\makeatletter

\newcommand*{\addFileDependency}[1]{
\typeout{(#1)}
\@addtofilelist{#1}
\IfFileExists{#1}{}{\typeout{No file #1.}}
}\makeatother

\newcommand*{\myexternaldocument}[1]{%
\externaldocument{#1}%
\addFileDependency{#1.tex}%
\addFileDependency{#1.aux}%
}

\myexternaldocument{Supp}

\usepackage[pagebackref,breaklinks,colorlinks]{hyperref}

\usepackage[capitalize]{cleveref}
\crefname{section}{Sec.}{Secs.}
\Crefname{section}{Section}{Sections}
\Crefname{table}{Table}{Tables}
\crefname{table}{Tab.}{Tabs.}

\def\cvprPaperID{84} 
\def\confName{CVPR}
\def\confYear{2023}

\begin{document}

\title{Efficient Semantic Segmentation by Altering Resolutions for Compressed Videos}

\author{Yubin Hu\textsuperscript{\rm 1} \:
    Yuze He\textsuperscript{\rm 1} \:
    Yanghao Li\textsuperscript{\rm 1} \:
    Jisheng Li\textsuperscript{\rm 1} \:
    Yuxing Han\textsuperscript{\rm 2} \:
    Jiangtao Wen\textsuperscript{\rm 3} \:
    Yong-Jin Liu\textsuperscript{\rm 1}\thanks{Corresponding author.}\\
    \textsuperscript{\rm 1}Department of Computer Science and Technology, Tsinghua University \\
    \textsuperscript{\rm 2}Shenzhen International Graduate School, Tsinghua University  \\
    \textsuperscript{\rm 3}Eastern Institute for Advanced Study\\
    \tt\small \{huyb20, hyz22, liyangha18\}@mails.tsinghua.edu.cn, jas0n1ee@icloud.com, \\
    \tt\small yuxinghan@sz.tsinghua.edu.cn, jtwen@eias.ac.cn, liuyongjin@tsinghua.edu.cn
}

\maketitle

\begin{abstract}
Video semantic segmentation (VSS) is a computationally expensive task due to the per-frame prediction for videos of high frame rates. In recent work, compact models or adaptive network strategies have been proposed for efficient VSS. However, they did not consider a crucial factor that affects the computational cost from the input side: \textbf{the input resolution}. In this paper, we propose an altering resolution framework called AR-Seg for compressed videos to achieve efficient VSS.  
AR-Seg aims to reduce the computational cost by using low resolution for non-keyframes. To prevent the performance degradation caused by downsampling, we design a Cross Resolution Feature Fusion (CReFF) module, and supervise it with a novel Feature Similarity Training (FST) strategy. Specifically, CReFF first makes use of motion vectors stored in a compressed video to warp features from high-resolution keyframes to low-resolution non-keyframes for better spatial alignment, and then selectively aggregates the warped features with local attention mechanism. Furthermore, the proposed FST supervises the aggregated features with high-resolution features through an explicit similarity loss and an implicit constraint from the shared decoding layer. 
Extensive experiments on CamVid and Cityscapes show that AR-Seg achieves state-of-the-art performance and is compatible with different segmentation backbones. On CamVid, AR-Seg saves 67\% computational cost (measured in GFLOPs) with the PSPNet18 backbone while maintaining high segmentation accuracy. Code:  \url{https://github.com/THU-LYJ-Lab/AR-Seg}.

\end{abstract}

\section{Introduction}
\label{sec:intro}

Video semantic segmentation (VSS) aims to predict pixel-wise semantic labels for each frame in a video sequence. In contrast to a single image, a video sequence is a series of consecutive image frames recorded at a certain frame rate (usually 25fps or higher). Applying image-based segmentation methods \cite{chen2017rethinking,long2015fully,xie2021segformer,yu2018bisenet,zhao2017pyramid} to a video
frame by frame consumes considerable computational resources.
To improve the efficiency of VSS, existing methods mainly focus on the design of network architectures. A class of methods proposes compact and efficient image-based architectures to reduce the computational overhead per-frame \cite{li2019dfanet,li2019partial,mehta2018espnet,yu2020bisenet,yu2018bisenet,zhao2018icnet,xie21segformer}. Another class of methods applies a deep model to keyframes and a shallow network for non-keyframes to avoid the repetitive computation \cite{jain2019accel,li2018low,mahasseni2017budget,Paul0GT20} for videos. 

\begin{figure}[t]
\centering
     \begin{subfigure}[b]{0.22\linewidth}
         \centering
         \includegraphics[width=1.0\linewidth]{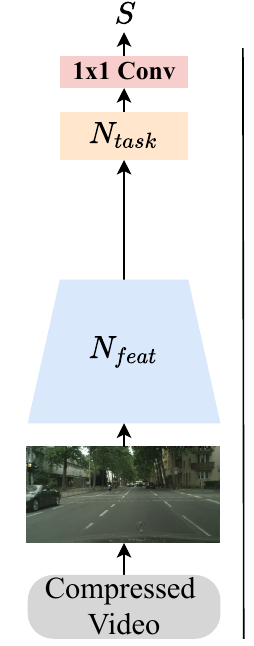}
         \caption{}
         \label{fig:per-frame}
     \end{subfigure}
     \hfill
     \begin{subfigure}[b]{0.39\linewidth}
         \centering
         \includegraphics[width=1.0\linewidth]{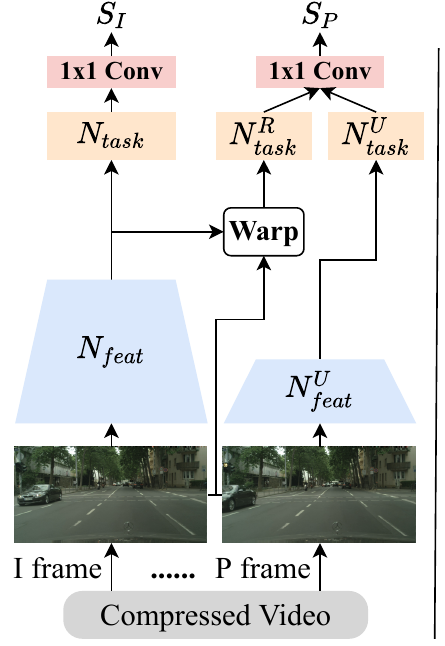}
         \caption{}
         \label{fig:accel}
     \end{subfigure}
     \hfill
     \begin{subfigure}[b]{0.37\linewidth}
         \centering
         \includegraphics[width=1.0\linewidth]{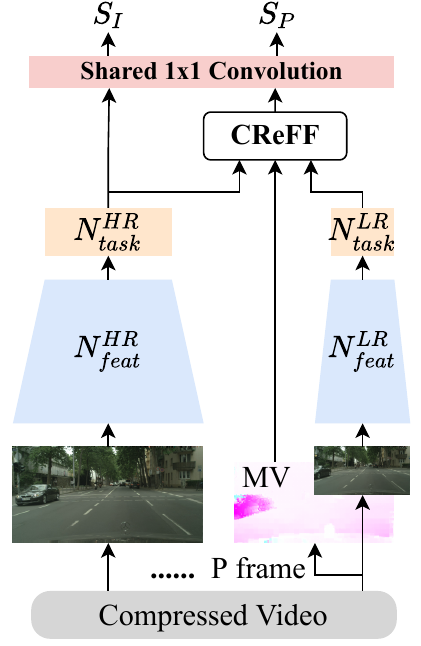}
         \caption{}
         \label{fig:ar-seg}
     \end{subfigure}

\caption{Comparison of different VSS methods: (a) per-frame framework, (b) Accel \cite{jain2019accel} that alters the depth of models, and (c) our AR-Seg. AR-Seg reduces the computational cost for non-keyframes by lowering the input resolution (depicted by narrow blocks), which is a dimension orthogonal to the depth of networks.
}
\label{fig:pipeline}
\vspace{-3mm}
\end{figure}

The work presented in this paper is based on an important observation: the above existing works ignored a crucial factor that affects the computational cost from the input side: \textbf{the input resolution}. For image-related tasks, the input resolution directly determines the amount of computation, e.g., the computational cost of 2D convolution is proportional to the product of image width and height. Once we downsample the input frame by $0.5\times0.5$, the computational overhead can be reduced by 75\%. On the other hand, decreasing resolution often leads to worse segmentation accuracy due to the loss of information \cite{wang2020dual,zhao2018icnet}. 
In this paper, we propose to prevent the accuracy degradation by using temporal correlation in the video.
Since the contents of video frames are temporally correlated, the local features lacking in low-resolution (LR) frames can be inferred and enriched by finding correspondences in sparse high-resolution (HR) reference frames based on motion cues. In a compressed video, the motion vectors contain such motion cues and can be obtained along with the video frames from video decoding at almost no additional cost.

Motivated by the above observation, we propose an altering resolution framework for compressed videos, named AR-Seg, to achieve efficient VSS. As shown in Figure \ref{fig:pipeline}\subref{fig:ar-seg}, AR-Seg uses an HR branch to process keyframes at high resolution and an LR branch to process non-keyframes at low resolution. 
In particular, to prevent performance drop caused by downsampling, we insert a novel Cross Resolution Feature Fusion (CReFF) module into the LR branch and supervise the training with a Feature Similarity Training (FST) strategy to enrich local details in the LR features.
CReFF fuses the HR keyframe features into LR non-keyframe features in two steps: 1) Align the spatial structures of features from different frames by feature warping with motion vectors, which can be readily obtained from compressed videos at almost no additional cost; 2) Selectively aggregate the warped features (which may be noisy after warping) into LR features with the aid of local attention mechanism. Since local attention assigns different importance to each 
location in the neighborhood, it is an effective way to avoid misleading by noisy warped features.

Furthermore, our proposed FST strategy guides the learning of the CReFF aggregated features.
FST consists of an \textit{explicit} similarity loss (between the aggregated features and HR features inferred from non-keyframes) and an \textit{implicit} constraint from the shared decoding layer across the HR and LR branches. Such a training strategy helps the LR branch to learn from features extracted from the HR branch, which is reliable and effective. Integrated with CReFF and FST, AR-Seg efficiently compensates for the accuracy loss of LR frames.
Overall, AR-Seg significantly reduces the computational cost of VSS by altering input resolutions, while maintaining high segmentation accuracy.

To sum up, we make three contributions in this paper: \textbf{1)} We propose an efficient framework for compressed videos, named AR-Seg, that uses altering input resolution for VSS and significantly reduces the computational cost without losing segmentation accuracy. \textbf{2)} We design an efficient CReFF module to prevent the accuracy loss by aggregating HR keyframe features into LR non-keyframe features. \textbf{3)} We propose a novel FST strategy that supervises the LR branch to learn from the HR branch through both explicit and implicit constraints. Experiment results demonstrate the effectiveness of AR-Seg with different resolutions, backbones, and keyframe intervals. On both CamVid \cite{brostow2009semantic} and Cityscapes \cite{cordts2016cityscapes} datasets, compared to the constant-resolution baselines, AR-Seg reduces the computational cost by nearly 70\% without compromising accuracy. 

\section{Related Works}
\label{section: related work}

As a fundamental task of scene understanding, semantic segmentation has been an active research area for many years \cite{ciresan2012deep,farabet2012learning,ladicky2009associative,tighe2010superparsing}, which also attracts considerable attention in the study of deep neural networks, e.g., FCN \cite{long2015fully}, DeepLabs \cite{chen2014semantic,chen2017deeplab} and PSPNet \cite{zhao2017pyramid}. In order to obtain accurate results in real-time applications, several methods have been proposed to improve the efficiency of semantic segmentation, which we summarize as follows. 

\textbf{Efficient Image Segmentation Methods.} Many compact architectures have been proposed for efficient image segmentation. DFANet \cite{li2019dfanet} adopted
a lightweight backbone to reduce computational cost and designed cross-level aggregation for feature refinement. DFNet \cite{li2019partial} utilized a partial order pruning
algorithm to search segmentation models for a good trade-off between speed and accuracy. ICNet \cite{zhao2018icnet} used a cascade fusion module and transformed part of the computation from high-resolution to low-resolution. Wang et al. \cite{wang2020dual} designed super-resolution learning to improve image segmentation performance. BiSeNets \cite{fan2021rethinking,yu2020bisenet,yu2018bisenet}
used two-stream paths for low-level details and high-level context information, respectively. ESPNet \cite{mehta2018espnet} used an efficient spatial pyramid to accelerate the convolution computation. These efficient backbone networks reduce the computational burden of single-image segmentation, and can be applied to temporal or spatial frameworks in VSS.

\textbf{Temporally Correlated Video Segmentation.} Another group of methods focus on utilizing temporal redundancy in videos. They proposed various mechanisms that propagate the deep features extracted from keyframes to reduce the computation for non-keyframes. Clockwork \cite{shelhamer2016clockwork} directly reused the segmentation result from keyframes, while Mahasseni et al. \cite{mahasseni2017budget}
interpolated segmentation results in the neighborhood. Noticing the lack of information from non-keyframes, Li et al. \cite{li2018low} extracted shallow features from non-keyframes, and fused them into the propagated deep features by spatially variant convolution. 
To compensate for the spatial misalignment between video frames, Zhu et al. \cite{zhu2017deep} and Xu et al. \cite{xu2018dynamic} warped the intermediate features from keyframes by optical flow to produce segmentation results for non-keyframes. Jain et al. \cite{jain2019accel} fused the shallow features of non-keyframe into the warped features, and decoded them into better results. With global attention mechanism, TD-Net \cite{hu2020temporally} aggregated the features from different time stamps and replaced the deep model with several shallow models distributed across the timeline. All the above methods mainly reduced the depth of backbone networks, but neglected the factor of input resolution considered in this paper. 
Instead of processing the image frames as a whole, Verelst et al. \cite{verelst2021blockcopy} split the frame into blocks and chose to copy or process them by a policy network. This block-based method reduces computational overhead from the spatial dimension, but lacks global information on non-keyframes. Kim et al. \cite{kim2020highway} 
attempted to improve efficiency by reducing resolution. But they directly used the LR segmentation results, thus suffering from severe performance degradation. Compared to these methods, our proposed AR-Seg keeps the global information of LR non-keyframes, and enhances LR frames by selectively aggregating intermediate features from HR keyframes. 

\textbf{Compressed-Domain Video Analysis.} Compressed video formats have been recently utilized in computer vision tasks. The motion vectors and residual maps are treated as additional modalities and directly fed into networks for video action recognition \cite{battash2020mimic,huo2020lightweight,shou2019dmc,wu2018compressed} and semantic segmentation \cite{tan2020real}.
Such motion information also helps compensate for the spatial misalignment of features from different frames. Wang et al. \cite{wang2019fast} leveraged motion vectors to warp features in previous frames for object detection. Fan et al. \cite{fan2021motion} conditionally executed the backbone network for pose estimation depending on the residual values. For VSS, several methods have been proposed for efficient segmentation in the compressed domain. Jain et al. \cite{jain2018fast} warped the former and latter keyframe features using motion vectors, and predicted the non-keyframe features by interpolation. Tanujaya et al. \cite{tanujaya2020semantic} warped the results of keyframe segmentation for non-keyframes, and refined the warped segmentation by guided inpainting. Feng et al. \cite{feng2020taplab} replaced a block of warped features with a local non-keyframe feature patch for further refinement. These methods reduced the computational cost of VSS, but suffered from performance degradation due to the limited capability of their feature refinement modules designed for non-keyframes.
In our proposed AR-Seg, the warped features are refined by local attention mechanism according to the LR features of non-keyframes. Our attention-based refinement selectively aggregates the warped features and effectively suppresses the noise in motion vectors, achieving good segmentation accuracy with little computational overhead.

\section{Method}

\begin{figure*}[h]
\centering
\includegraphics[width=0.95\linewidth]{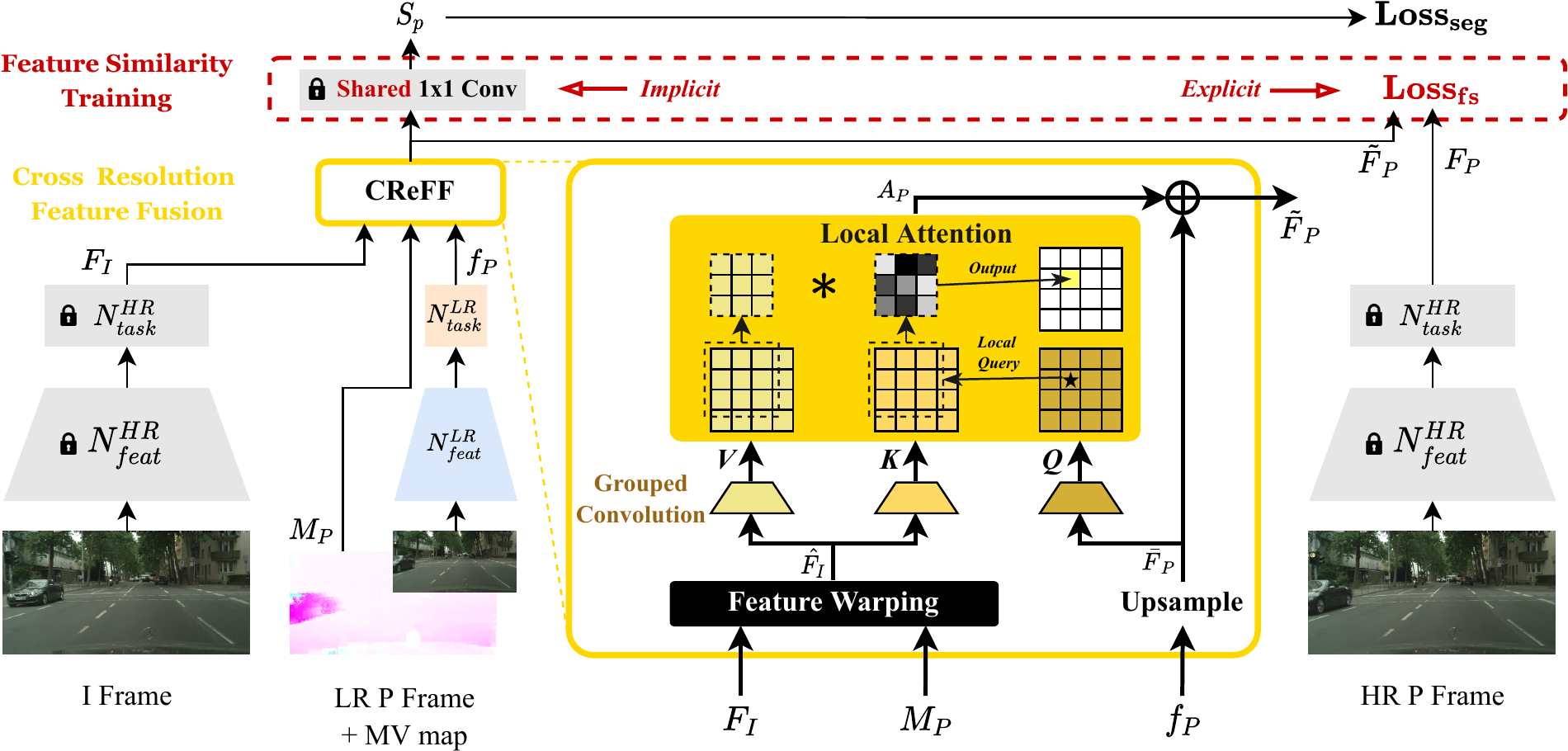}
\caption{The CReFF module in the network architecture and feature similarity training (FST) strategy. CReFF consists of feature warping $\mathcal{W}_{MV}$ and local-attention-based feature fusion $\mathcal{F}_{LA}$ (Section \ref{section: CREFF}). FST includes the explicit supervision by the feature similarity loss and the implicit supervision from shared convolution (Section \ref{sec:FST}). Parameters in gray blocks are \textbf{fixed} during the training of the LR branch.}
\label{fig:detail}
\vspace{-3mm}
\end{figure*}

In order to achieve efficient VSS for compressed videos, we propose an altering resolution framework named AR-Seg (Section \ref{sec:framework}). AR-Seg uses two branches to process HR keyframes and LR non-keyframes in the video separately. To compensate for the information loss due to downsampling the non-keyframes, we design a novel CReFF module that aggregates HR keyframe features into LR non-keyframe features (Section \ref{section: CREFF}). To guide the learning of aggregated features, we further propose a novel feature similarity training (FST) strategy containing both explicit and implicit constraints (Section \ref{sec:FST}).

\subsection{AR-Seg Framework}
\label{sec:framework}

For image/video semantic segmentation tasks, the input resolution directly determines the amount of computation, no matter what type of algorithms are applied, e.g., Conditional Random Fields, CNNs, and Transformers. Although reducing resolution has been studied in the context of video recognition \cite{Meng2020,Ma2022}, altering resolution in dense prediction tasks like VSS remains unexplored. Based on this observation, we design the AR-Seg framework for VSS, which inputs video frames with altering resolutions; i.e., in AR-Seg, only a few keyframes are processed at high resolution to preserve fine details, while other non-keyframes are processed at low resolution to reduce the computational cost. 

To identify the keyframes, we make use of the frame structure in a group of pictures (GOP) encoded in compressed videos \cite{sullivan2012overview,wiegand2003overview}. A GOP includes $L$ consecutive frames of three types: I frame, P frame, and B frame. 
I frames are encoded in intra mode, while P and B frames are encoded in inter mode that computes motion vectors for motion compensation. 
In each GOP, we treat the first I frame as a keyframe and process it at high resolution. The remaining $L-1$ frames in GOP are non-keyframes and processed at low resolution. To simplify the description of our method, following the previous works \cite{feng2020taplab,wu2018compressed}, we only consider the GOP structure without B frames in this section. The full treatment including B frames is presented in Appendix A4.

Due to the domain gap between images with different resolutions, it is difficult for a single network to extract features suitable for both HR and LR resolution images. Thus we design two branches with shared backbone architecture in the model and train them separately for each resolution. 
Figure \ref{fig:pipeline}\subref{fig:ar-seg}
summarizes the proposed AR-Seg framework. It consists of two branches: an HR branch for keyframes and an LR branch for non-keyframes. 

The HR branch adapts an image segmentation network, which consists of a feature sub-network $N_{feat}$, a task sub-network $N_{task}$, and a final 1x1 convolutional layer\footnote{We follow the modular division of backbone networks in \cite{jain2019accel}. Two backbones are discussed and compared in Section \ref{sec:experiment}.}. It predicts segmentation results in high-resolution and simultaneously provides intermediate features before the final convolution as a reference for the LR branch. 
The LR branch is equipped with the same backbone network as the HR branch. To prevent performance degradation caused by downsampling, we design a CReFF module that aggregates HR features of the keyframe into the LR branch and place it before the final convolution layer of the backbone network. 
CReFF aggregates the HR reference features into the extracted LR features, yielding estimated HR features for the non-keyframes, which are further converted into pixel-wise semantic labels by the final convolution layer.

As illustrated in Figure \ref{fig:pipeline},
different from the previous Accel framework \cite{jain2019accel}, AR-Seg performs feature fusion before the last 1x1 convolution layer, instead of before the task sub-network $N_{task}$. The reason is twofold: 1) Since feature maps before the final convolution have basically the same spatial layout as the input images and segmentation outputs, we can utilize motion vectors to compensate for the spatial misalignment of features at such position; and 
2) As the CReFF module actually upsamples the LR features, such a placement allows almost all convolution layers to benefit from the low resolution, thus reducing most of the computational cost.

\subsection{CReFF: \ Cross Resolution Feature Fusion}
\label{section: CREFF}
In the AR-Seg framework, CReFF aims to prevent the performance degradation caused by the lack of fine local details in LR non-keyframes.
Unlike a single image, video frames are intrinsically temporally correlated, so missing details in LR non-keyframes can be retrieved from the corresponding regions in HR keyframes according to motion cues. Motion vectors (MVs) in the compressed video exactly provide such motion cues at the block level, i.e., pixels inside a macroblock share the same motion vector. Almost all mainstream video compression standards use motion vectors for inter-prediction, including H.26x series \cite{sullivan2012overview,wiegand2003overview}, AOMedia series \cite{chen2018overview} and AVS series \cite{Liang2004avs,gao2014overview,zhang2019recent}. Such block-wise MVs are readily available in compressed videos and can be used to assist the LR branch.

Specifically, as depicted in Figure \ref{fig:detail}, the HR branch of AR-Seg extracts the feature $F_{I} \in \mathcal{R}^{C\times H \times W}$ from an I frame, and the LR branch extracts the feature $f_{P} \in \mathcal{R}^{C\times h \times w}$ from a P frame. 
Although P frames are processed in low resolution, CReFF takes $F_I$, $M_P$, and $f_P$ as input to generate the aggregated feature $\tilde{F}_P$, where $M_P \in \mathcal{R}^{2\times H \times W}$ denotes the MVs from P frame to I frame. The two channels of $M_P$ correspond to $x$ and $y$ dimensions of motion vectors, denoted by $c_x$ and $c_y$.
Inside the CReFF module, the MV-based feature warping operation $\mathcal{W}_{MV}$ firstly warps $F_I$ to the spatial layout of the P frame, which can be formulated as per-pixel shifting: 
\begin{equation}
    \hat{F}_I^{(x,y)} = F_I^{(x+M_P^{(c_x,x,y)},y+M_P^{(c_y,x,y)})},
\end{equation}
where $\hat{F}_I \in \mathcal{R}^{C\times H \times W}$ denotes the warped HR feature that will be further fused into LR features. 

Due to the coarse-grained MVs (block-level instead of pixel-level) and the varying occlusion relationships across video frames, the warped features $\hat{F}_I$ are often noisy and misleading. 
Inspired by the success of non-local operation \cite{wang2018non} and attention mechanism \cite{hu2020temporally,Oh2019video,PaulDGT21Local} in video-based applications, we propose to assign different fusion importance weights to the $(x,y)$ locations in noisy features $\hat{F}_I$ by attention mechanism.
Since $\hat{F}_I$ is roughly spatially aligned to $f_p$ after warping, we use local attention to efficiently fuse the features as follows. 

In the local-attention-based feature fusion module $\mathcal{F}_{LA}$, 
we firstly generate the \textbf{\textit{Value}} and \textbf{\textit{Key}} feature maps from the warped HR features $\hat{F}_I$, and the \textbf{\textit{Query}} maps from the upsampled LR features $\bar{F}_P$. The $3 \times 3$ grouped convolution $\mathbf{Conv}_g$ with $groups=C$ is selected to efficiently encode the feature maps into attention representations $V_I , K_I, Q_P \in \mathcal{R}^{C\times H \times W}$. Note that attention representations share the same channel size as the intermediate features.
Denote the $n\times n$ neighborhood centered at $(x,y)$ as $Nbhd_{(x,y)}$. Within $Nbhd_{(x,y)}$, the output of local attention $A_P$ at the position $(x,y)$ is generated by 
\begin{equation}
    {A_P}^{(x,y)} = \overline{{V_I}^{Nbhd_{(x,y)}}} \mathcal{S} (\overline{{K_I}^{Nbhd_{(x,y)}}}, {Q_P}^{(x,y)}),
\end{equation}
where ${A_P}^{(x,y)},\ {Q_P}^{(x,y)} \in \mathcal{R}^{C\times 1}$ denote the feature vectors at the position $(x,y)$ of $A_P$ and $Q_P$ respectively, and $\overline{{V_I}^{Nbhd_{(x,y)}}},\ \overline{{K_I}^{Nbhd_{(x,y)}}} \in \mathcal{R}^{C \times n^2}$ are the re-arranged feature vectors within ${Nbhd_{(x,y)}}$ in $V_I$ and $K_I$, respectively. The similarity operation $\mathcal{S}(K, Q)$ is formulated as 
\begin{equation}
    \mathcal{S}(K, Q) = \mathit{Softmax} (\frac{K^T Q}{\sqrt{C}}).
\end{equation}
Furthermore, the aggregated feature $\tilde{F}_P$ for P frame is obtained in a residual fashion:
\begin{equation}
    \tilde{F}_P = \bar{F}_P + A_P = \mathit{Upsample}(f_P) + A_P.
\end{equation}

In summary, using the CReFF module, the feature details from the I frame are firstly aligned to the P frame, and then aggregated into the LR branch according to the pixel-wise similarity between $Q_P$ and $K_I$. The verification of the architecture design of CReFF is presented in Section \ref{sec:ablation}. The reader is referred to Appendix A1 for more details on the visualization of attention weights in $\mathcal{F}_{LA}$.

\subsection{FST: \ Feature Similarity Training}
\label{sec:FST}

In order to effectively train the CReFF module, we propose a feature similarity training (FST) strategy. FST utilizes the HR features of P frame $F_P$ (extracted from the HR branch) to guide the learning of the aggregated features $\tilde{F}_P$ in the LR branch.
Since $F_P$ contains sufficient details to produce high-quality segmentation results, CReFF can learn how to aggregate $\bar{F}_P$ and $\hat{F}_I$ into effective HR features from it under the supervision of FST. 
Specifically, FST supervises the training process of the LR branch both \textit{explicitly} and \textit{implicitly} in the following ways.

The \textit{explicit} constraint is to use the feature similarity loss $\mathcal{L}_{fs}$. We use mean square error ($MSE$) to measure the difference between $\tilde{F}_P$ and $F_P$, which serves as an additional regularization for the LR model:
\begin{equation}
            \mathcal{L} = \mathcal{L}_{seg} + \mathcal{L}_{fs} = {CE}({S}_{P}, G_{P}) + {MSE}(\tilde{F}_{P},F_{P}),
\label{eq: loss}
\end{equation}
where ${S}_{P} \in \mathcal{R}^{C_{out} \times H \times W}$ denotes the segmentation result produced by LR branch, $G_P \in \mathcal{R}^{H \times W}$ denotes the ground-truth segmentation of P frame and the segmentation loss $\mathcal{L}_{seg}$ is the standard cross entropy loss ${CE}({S}_{P},G_{P})$. 

The \textit{implicit} constraint of FST is the shared decoding layer of $\tilde{F}_P$ and $F_P$. In the segmentation backbone model trained on the HR images, the final convolution layer acts as the segmentation decoder, which contains deep semantic information about high-quality HR features. To utilize such information, we directly transfer the final $1\times1$ convolution layer of the HR branch to the LR branch with fixed parameters. Since the parameters are trained on HR features, they produce better segmentation results $S_P$ when $\tilde{F}_P$ is closer to the HR feature $F_P$. 


In summary, with the \textit{explicit} and \textit{implicit} constraints, FST effectively transfers the knowledge of HR features from the HR branch to the LR branch, enabling high-quality segmentation based on the aggregated features of CReFF.
Figure \ref{fig:detail} shows the overall training strategy for the LR branch. The HR I frame provides the features $F_{I}$ for feature fusion in CReFF, and the HR P frame provides the features $F_P$ for the \textit{explicit} supervision in FST. Parameters of the LR branch are trained using the total loss $\mathcal{L}$ via backpropagation, with fixed parameters of the HR branch and the shared final convolution layer. 


\section{Experiments}
\label{sec:experiment}

We evaluate the proposed AR-Seg framework on CamVid \cite{brostow2009semantic} and Cityscapes \cite{cordts2016cityscapes} datasets for street-view video semantic segmentation. Below we present experiments to demonstrate the efficiency of AR-Seg and its compatibility with different backbone models, resolutions, and video compression configurations.

\subsection{Experimental Setup}

\noindent \textbf{Datasets \& Pre-processing.} 
The \textit{CamVid} \cite{brostow2009semantic} dataset consists of 4 videos of $720\times960$ resolution captured at 30 fps, with semantic annotations at 1Hz and, in part, 15Hz. The \textit{Cityscapes} \cite{cordts2016cityscapes} dataset contains street view videos of $1024\times2048$ resolution captured in 17 fps, from which 5,000 images are densely annotated. We use the official train/validation/test split for both datasets. Following the previous works \cite{hu2020temporally,yu2020bisenet}, we select 11 and 19 classes for training and evaluation on these two datasets, respectively.

To simulate a real video compressed scenario, we compress the videos at reasonable bit-rates of 3Mbps for CamVid and 5Mbps for Cityscapes with the HEVC/H.265 \cite{sullivan2012overview} standard. The reader is referred to Appendix A3.1 for the detailed pre-processing steps.


\noindent \textbf{Models \& Baselines.} 
To demonstrate the compatibility of AR-Seg with different backbones, we select two representative image segmentation models in our experiments: PSPNet \cite{zhao2017pyramid} and BiseNet \cite{yu2018bisenet}, which is similar to settings in the previous work \cite{hu2020temporally}. The former is a widely used classical model, and the latter is a lightweight model that achieves state-of-the-art performance.
We use \textbf{AR$^\alpha$-} as the prefix of AR-Seg with specified backbone networks, where $\alpha$ denotes the downsample scale for the LR branch. 

\noindent \textbf{Training \& Evaluation Details.} 
Given a GOP length $L$, we train the LR branch with image pairs $(i, p)$, where $p$ refers to the P frame with annotation and $i = p-(L-1)$ refers to the I frame as a reference. We denote the distance between 
the annotated and the reference frames as $d$, then $d=L-1$ for the training pairs. 

For evaluation, we test AR-Seg with different distances $d$ between the target frame $p$ and the reference keyframe $i$. For $d = 0$, we treat frame $p$ as the keyframe and process it by the HR branch. Otherwise, we feed frame $p$ into the LR branch for $d \in (0,L-1]$. The average of $\mathit{mIoU_d}$ for each distance $d$ is reported as the mIoU result. 
We measure FLOPs by PyTorch-OpCounter \cite{PyTorch-OpCounter} following the previous methods \cite{nirkin2021hyperseg,orsic2019defense}. 
All the comparisons are evaluated on the compressed videos. More training and evaluation details are presented in  Appendix A3.2 and A3.3.

\begin{figure*}[t]
\centering
\includegraphics[width=1.0\linewidth]{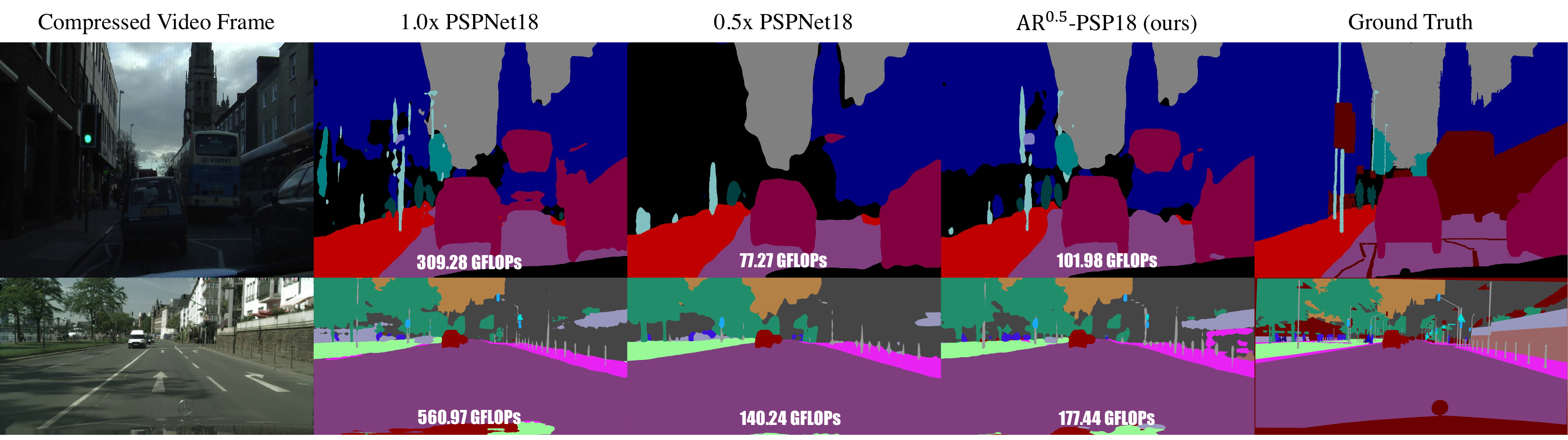}
\caption{Semantic segmentation on CamVid (top) and Cityscapes (bottom) with $d=11$. Note that AR$^{0.5}$-PSP18 predicts more semantic details than the constant-resolution PSPNet18 working on 0.5x resolution. Compared to the 1.0x baseline, AR$^{0.5}$-PSP18 generates similar segmentation results, but consumes only 33.0\% computational cost (measured in GFLOPs). }
\label{fig:subject}
\end{figure*}

\subsection{Experiment Results}

\noindent\textbf{Comparison with image-based methods.} We first compare AR-Segs (with PSPNet \cite{zhao2017pyramid} and BiseNet \cite{yu2018bisenet} as backbone) to their image-based counterparts of 1.0x resolution. 
As shown in Table \ref{tab:image_base}, on both CamVid and Cityscapes datasets, the proposed AR$^{0.5}$- models achieve on-par or better performance than the 1.0x resolution baselines while saving ~67\% computational cost. Different from the low-resolution baselines that lead to significant performance degradation, AR-Seg successfully preserves the segmentation accuracy with the help of the CReFF module and the FST strategy. More comparisons between AR-Seg and the LR baselines under different resolutions are presented in Appendix A2.1.
Furthermore, these experiment results with PSPNet and BiseNet also demonstrate the compatibility of AR-Seg for different backbone networks. 

\noindent\textbf{Comparison with video-based methods.} Taking temporal coherence into consideration, we compare AR-Seg with the recent state-of-the-art video-based methods for efficient VSS. Besides the accuracy and computational cost, we also follow the previous work \cite{PaulDGT21Local} to report the relative changes compared to their single-frame backbone models. Specifically, $\widetilde{\Delta}$mIoU denotes the relative change of mIoU, and $\widetilde{\Delta}$GFLOPs denotes the relative change of GFLOPs. As shown in Table \ref{tab:video_base}, existing video-based methods usually improve accuracy ($\widetilde{\Delta}$mIoU$>$0) at the cost of more computation ($\widetilde{\Delta}$GFLOPs$>$0), e.g., TDNet \cite{hu2020temporally} and Accel \cite{jain2019accel}. Other methods, including BlockCopy \cite{verelst2021blockcopy}, TapLab \cite{feng2020taplab} and Jain et al. \cite{jain2018fast}, reduce the computational cost ($\widetilde{\Delta}$GFLOPs$<$0) but the accuracy also decreases 3\%-7\%  ($\widetilde{\Delta}$mIoU$<$0). As a comparison, our proposed models AR$^{0.6}$-  can reduce the computational cost (\textbf{$\widetilde{\Delta}$GFLOPs$<$0}) by more than 55\% and preserve the accuracy of single-frame backbone models (\textbf{$\widetilde{\Delta}$mIoU$\geq$0}). With the lightweight backbone model BiseNet, AR$^{0.6}$-Bise18 achieves good performance in both accuracy and computational cost. More results of video-based methods and their single-frame backbone models are presented in Appendix A2.2.

\begin{table}[t]
    \small
    \centering
    \caption{Comparison to the image-based methods on CamVid \textit{test} set and Cityscapes \textit{valid} set.}
    \vspace{-3mm}
    \begingroup
    \setlength{\tabcolsep}{3.5pt}
    \begin{tabular}{l | c |c c|c c}
        \specialrule{1.5pt}{1pt}{1pt}
         & \multirow{2}{*}{Method}
         &  \multicolumn{2}{|c|}{PSPNet18\cite{zhao2017pyramid}} 
         &  \multicolumn{2}{c}{BiseNet18\cite{yu2018bisenet}} 
         \\
        \cline{3-6}
         & & mIoU(\%)$\uparrow$ & GFLOPs $\downarrow$ & mIoU(\%)$\uparrow$ & GFLOPs $\downarrow$ \\
        \toprule
        \multirow{4}{*}{\rotatebox{90}{CamVid}} & 1.0x & 69.43 & 309.02 & 71.57 & 58.83 \\
        \cline{2-6}
        & AR$^{0.7}$ & \textbf{71.23} & 169.86 & \textbf{71.78} & 31.89 \\
        & AR$^{0.6}$ & 70.82 & 133.09 & 71.60 & 24.68 \\
        & AR$^{0.5}$ & 70.48 & \textbf{101.98} & 70.38 & \textbf{18.96} \\
        \toprule
        \multirow{4}{*}{\rotatebox{90}{Cityscapes}} & 1.0x & 69.00 & 560.97 & 70.09 & 178.96\\
        \cline{2-6}
        & AR$^{0.7}$ & \textbf{70.23} & 302.95 & \textbf{70.86} & 97.10\\
        & AR$^{0.6}$ & 69.45 & 234.91 & 70.72 & 76.06\\
        & AR$^{0.5}$ &  69.03 & \textbf{177.44} & 70.57 & \textbf{57.00} \\
        \specialrule{1.5pt}{1pt}{1pt}
    \end{tabular}
    \endgroup
    \label{tab:image_base}
\end{table}

\begin{table}[t]
\centering
\caption{Comparison to the video-based methods on CamVid \textit{test} set and Cityscapes \textit{valid} set. $\widetilde{\Delta}x=\frac{\Delta x}{|x|}$ denotes the relative change compared to their single-frame backbone models. The best results are bold and the second best results are underlined.}
\vspace{-3mm}
\label{tab:video_base}
\small
\begingroup
\setlength{\tabcolsep}{2pt}
\begin{tabular}{l | l | c c | c c}
\specialrule{1.5pt}{1pt}{1pt}
& Method & mIoU$\uparrow$ & GFLOPs$\downarrow$ & $\widetilde{\Delta}$mIoU$\uparrow$ & $\widetilde{\Delta}$GFLOPs$\downarrow$ \\
\toprule
\multirow{7}{*}{\rotatebox{90}{CamVid}} & Accel-DL18 \cite{jain2019accel} & 66.15 & 397.70 & \textbf{+13.8\%} & +61.9\%\\
& TD$^4$-PSP18 \cite{hu2020temporally} & 70.13 & 363.70 & +1.0\% & +17.7\%\\

& BlockCopy \cite{verelst2021blockcopy}  & 66.75  & 107.52 & -5.2\% & -45.7\% \\

& TapLab-BL2 \cite{feng2020taplab} & 67.57 & 117.73 & -3.1\% & -50.2\%\\

& Jain et al. \cite{jain2018fast} & 67.61 & 146.97 & -4.3\% & -53.8\%\\
\cline{2-6}

& AR$^{0.6}$-PSP18  & \underline{70.82} & \underline{101.98} & \underline{+2.0\%} & \underline{-57.0\%} \\
& AR$^{0.6}$-Bise18  &  \textbf{71.60} & \textbf{24.68} & +0.0\% & \textbf{-58.0\%}\\

\toprule
\multirow{7}{*}{\rotatebox{90}{Cityscapes}}& Accel-DL18 \cite{jain2019accel} & 68.25 & 1011.75 & \textbf{+18.4\%} & +96.0\% \\
& TD$^4$-PSP18 \cite{hu2020temporally} &  \underline{70.11} & 673.06 & \underline{+1.6\%} & +20.0\%\\
& BlockCopy \cite{verelst2021blockcopy}  &  67.69 & 294.20 & -6.7\% & -41.2\%\\
& TapLab-BL2 \cite{feng2020taplab} &  68.90 & 237.29 & -4.1\% & -50.6\%\\

& Jain et al. \cite{jain2018fast} & 68.57 & 342.67 & -5.1\% & -52.5\% \\
\cline{2-6}

& AR$^{0.6}$-PSP18 &  69.45 & \underline{234.91} & +0.7\% & \textbf{-58.1\%}\\
& AR$^{0.6}$-Bise18 &  \textbf{70.72} & \textbf{76.06} & +0.9\% & \underline{-57.5\%}\\
\specialrule{1.5pt}{1pt}{1pt}
\end{tabular}
\endgroup

\vspace{-3mm}
\end{table}

\subsection{Ablation Study}
\label{sec:ablation}

We perform ablation studies to show the importance of each component in CReFF and FST, as well as the location of CReFF. We also evaluate AR-Seg in terms of different resolutions and GOP lengths, which reflects the influence of hyper-parameters $\alpha$ and $L$, respectively.
We conducted ablation studies on the 30fps CamVid dataset, and used PSPNet18 as the default backbone model, $L=12$ as the default GOP length, and HEVC@3Mbps as the default codec. 

\noindent \textbf{Architecture of CReFF.}  We first validate the necessity of $\mathcal{W}_{MV}$ and $\mathcal{F}_{LA}$. The method without CReFF directly applies FST to the upsampled features $\bar{F}_P$ and serves as a baseline. As shown in Table \ref{tab:ablation}, simply warping the keyframe features (+ $\mathcal{W}_{MV}$) saves the most amount of computation by skipping processing non-keyframes with the segmentation network, but receives poor accuracy. Directly fusing the keyframe features using local attention (+ $\mathcal{F}_{LA}$) does not perform very well, because the feature maps are not well-aligned due to object motion in videos. 

We further evaluate the design of $\mathcal{F}_{LA}$ by replacing 7x7 local attention ($LA$) with other operations, including $LA$ with different neighborhood sizes, $LA$ with non-grouped convolution encoders ($\mathcal{F}_{LA^*}$), global attention ($\mathcal{F}_{GA}$) and one-layer convolution ($\mathcal{F}_{Conv}$). 
For $\mathcal{F}_{GA}$, we downsample the \textbf{\textit{Value}} and \textbf{\textit{Key}} maps by $\frac{1}{32}$ to save the computation. 
$\mathcal{F}_{Conv}$ processes the concatenated feature $[\hat{F}_I, \bar{F}_P]$ with a 3x3 convolution. Due to the large channel number in deep layers, such an operation brings considerable computation overhead.
Results in Table \ref{tab:ablation} show that $\mathcal{F}_{LA}$ with a 7x7 neighborhood achieves a good balance between computation and accuracy. Other designs increase the computational cost without improving the accuracy.
Furthermore, removing the direct connection (DC) path from CReFF reduces the mIoU to 69.14\%. This result implies that CReFF is more likely to learn the residuals of HR features than the absolute values. 


\begin{table}[t]
\small
\centering
\caption{Ablation experiments on CamVid dataset with PSPNet18. Settings used in our final model are underlined. }
\vspace{-3mm}
\label{tab:ablation}
\begingroup
\setlength{\tabcolsep}{3pt}
\begin{tabular}{l l c c }
\specialrule{1.5pt}{1pt}{1pt}
Experiment & Method & mIoU(\%) & GFLOPs \\
\toprule
\multirow{2}{0.16\linewidth}{Baseline} 
&PSPNet18 (1.0x) & 69.43 & 309.02 \\
&PSPNet18 (0.5x) & 66.51 & 77.27\\
\toprule
\multirow{9}{0.16\linewidth}{Architecture of CReFF} 
& \underline{+ $\mathcal{W}_{MV}$ + $\mathcal{F}_{LA}$ (7x7)}  &  \textbf{70.48} & \textbf{101.98} \\
& w/o CReFF & 67.14 & 96.60\\
& + $\mathcal{W}_{MV}$ & 57.64 & 25.75\\
& + $\mathcal{F}_{LA}$ (7x7) & 67.93 &  101.98 \\
& + $\mathcal{W}_{MV}$ + $\mathcal{F}_{LA}$ (3x3) & 70.30 & 98.74 \\
& + $\mathcal{W}_{MV}$ + $\mathcal{F}_{LA}$ (11x11)  & 70.48 & 107.32\\
& + $\mathcal{W}_{MV}$ + $\mathcal{F}_{LA^*}$ (7x7) & 69.99 & 170.96 \\
& + $\mathcal{W}_{MV}$ + $\mathcal{F}_{GA}$ (1/32) & 67.11 & 113.58\\
& + $\mathcal{W}_{MV}$ + $\mathcal{F}_{Conv}$ & 70.45 & 143.63\\
& + CReFF w/o DC  & 69.14 &  101.98 \\
\toprule
\multirow{3}{0.16\linewidth}{Location of CReFF} & before \underline{$C_{1\times1}$}   & \textbf{70.48} & \textbf{101.98} \\
& before $N_{task}$ & 68.60 & 214.76 \\
& before $N_{feat}$ & 68.31 & 308.46 \\

\toprule
\multirow{4}{0.16\linewidth}{Feature Similarity Training (FST)} & \underline{+ MSE Loss + Shared $C_{1\times1}$} & \textbf{70.48} & 101.98 \\
& w/o FST  & 69.21 & 101.98 \\
& + Shared $C_{1\times1}$ & 69.57 &  101.98 \\
& + MSE Loss & 70.17 & 101.98 \\
& + KL Loss + Shared $C_{1\times1}$ & 68.91 & 101.98 \\

\toprule
\multirow{4}{0.16\linewidth}{Keyframe Interval} & \underline{AR$^{0.5}$-PSP18, L=12}   & \textbf{70.48} & 101.98 \\
& AR$^{0.5}$-PSP18, L=15 & 70.28 & 97.88 \\
& AR$^{0.5}$-PSP18, L=20 & 70.28 & 94.11 \\
& AR$^{0.5}$-PSP18, L=30 & 69.67 & \textbf{90.34}\\

\specialrule{1.5pt}{1pt}{1pt}
\end{tabular}
\endgroup
\vspace{-3mm}
\end{table}

\noindent \textbf{Location of CReFF.} As specified in Section \ref{sec:framework}, we place CReFF before the final 1x1 convolution layer, which is different from the previous Accel framework \cite{jain2019accel}. To evaluate the influence of the split point location, we insert CReFF before different layers and evaluate the performance. Split points include the final convolution layer ($C_{1\times1}$), the task sub-network $N_{task}$ and the feature sub-network $N_{feat}$. For the $N_{feat}$ case, we insert CReFF after the first convolution layer of ResNet. 
As shown in Table \ref{tab:ablation}, placing CReFF before $C_{1\times1}$ results in the best performance, which affirms our design in Section \ref{sec:framework}. We note that fusing features at an early stage does not improve accuracy, but rather considerably increases computational cost.  

\noindent \textbf{Feature Similarity Training (FST).} The proposed FST strategy consists of the MSE Loss and the shared final convolution layer $C_{1\times1}$. We train AR$^{0.5}$-PSP18 with or without these components and report the results in Table \ref{tab:ablation}. Both components improve the segmentation performance compared to the model trained without FST. We also replace the MSE Loss with the Kullback-Leibler (KL) Divergence Loss, but the resulting segmentation performance is poor.

\begin{figure}[t]
\centering
     \begin{subfigure}[b]{0.595\linewidth}
         \centering
         \includegraphics[width=1.0\linewidth]{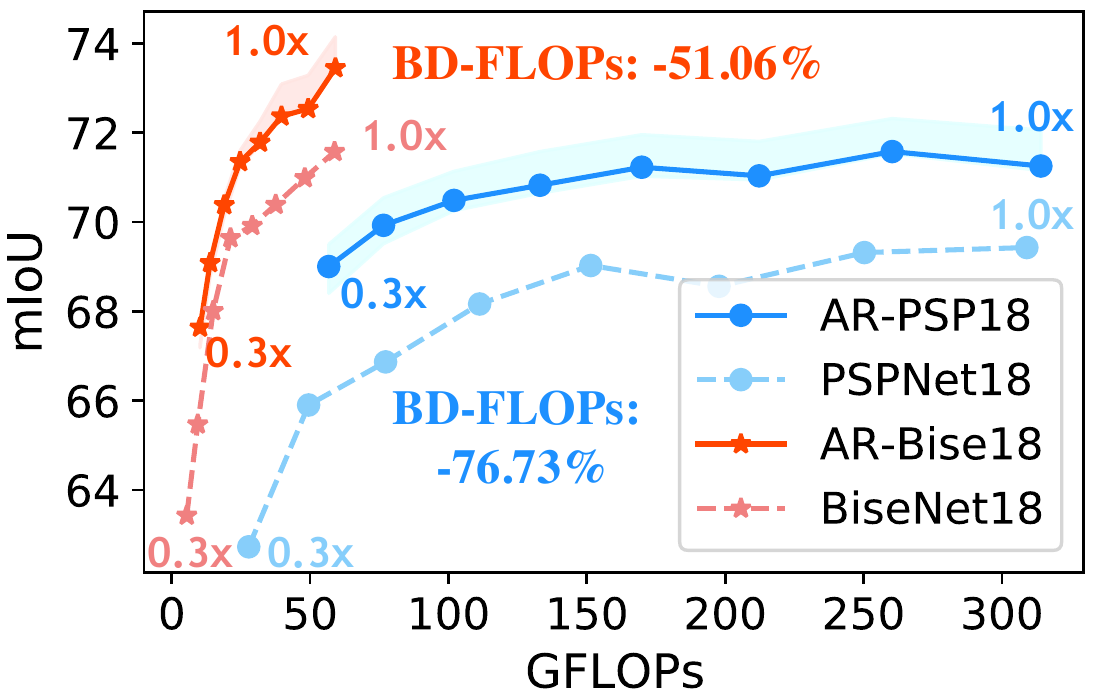}
         \caption{}
         \label{fig:resolution-table-1}
     \end{subfigure}
     \hfill
     \begin{subfigure}[b]{0.39\linewidth}
         \centering
         \includegraphics[width=1.0\linewidth]{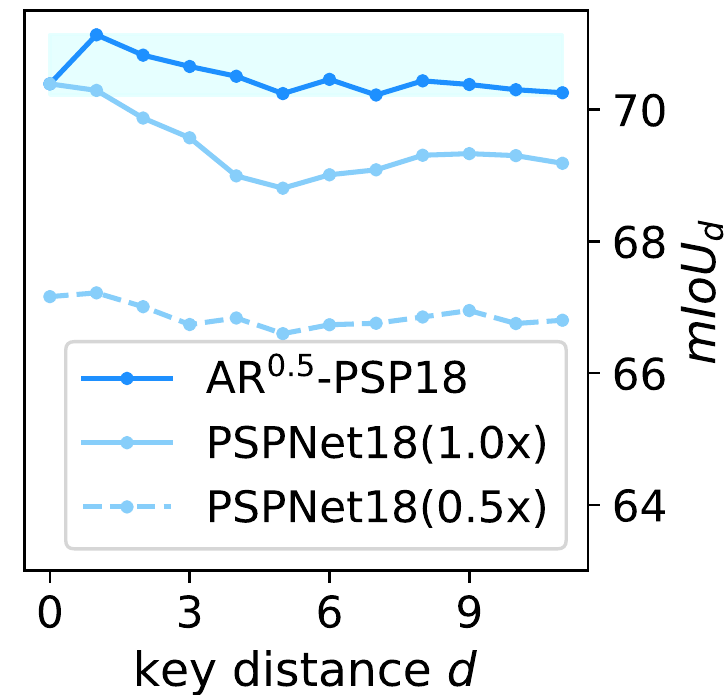}
         \caption{}
         \label{fig:resolution-table-2}
     \end{subfigure}
\vspace{-7mm}
\caption{(a) Performance of AR-Seg with different resolutions for the LR branch. (b) $\mathit{mIoU_d}$ for annotated frames with different distances to key frame. Value intervals of $\mathit{mIoU_d}$ when $d$ varies from $1$ to $L-1$ are depicted as color bars in (a).}
\vspace{-3mm}
\label{fig:resolution}
\end{figure}

\noindent \textbf{Resolution of LR Branch.} By adjusting the resolution of the LR branch, AR-Seg can tailor the setting adaptive to different computational budgets. Here, we train and evaluate AR-Seg with different resolutions for the LR branch, ranging from 0.3x to 1.0x. We also train and evaluate the constant-resolution baselines of each resolution for comparison. As shown in Figure \ref{fig:resolution}\subref{fig:resolution-table-1}, AR-Seg improves both backbones under all resolutions, demonstrating the effectiveness of CReFF and FST. To quantify the average improvement, we utilize two metrics \textit{BD-FLOPs} and \textit{BD-mIoU} following the design of BD-Rate and BD-PSNR \cite{bjontegaard2001calculation} in video compression. The results show that (1) with the same computational budget, AR-Seg improves the absolute accuracy for two backbones by 3.67\% and 3.02\% respectively, and (2) with the same accuracy, AR-Seg reduces the computational cost by 76.73\% and 51.06\% respectively. Both metrics are described in detail in Appendix A5.

\noindent \textbf{The Temporal Gap.} To investigate the influence of the distance $d$ to the keyframe, we plot the $\mathit{mIoU_d}$ results for AR$^{0.5}$-PSP18 and the constant-resolution baselines in Figure \ref{fig:resolution}\subref{fig:resolution-table-2}. As $mIoU_0$ is determined by the HR branch, AR$^{0.5}$-PSP18 shares the same point with PSPNet18(1.0x) at $d=0$. The $\mathit{mIoU_0}$ for PSPNet18(0.5x) is much lower due to downsampling.
When $d>0$, the accuracy of PSPNet18(1.0x) decreases since the compression artifacts in P frames are more severe than those in I frames. As a comparison, the AR$^{0.5}$-PSP18 benefits from the CReFF module and maintains high accuracy for all the $d$ values.

\noindent \textbf{Keyframe Intervals.} To validate the long-range reference, we extend our evaluation to different keyframe intervals without re-training. As shown in Table \ref{tab:ablation}, AR$^{0.5}$-PSP18 trained with $L=12$ maintains good performance with different GOP lengths. Moreover, even for $L=30$, which stands for 1s in 30fps videos and is larger than the discussion in previous works \cite{jain2019accel,zhu2017deep}, AR-Seg outperforms the 1.0x baseline using only 29.2\% FLOPs. 

\begin{table}[t]
\small
\centering
\caption{Performance of AR-Seg on videos compressed by different codecs. AR-Seg achieves comparable or even better accuracy than its image-based constant-resolution counterparts.}
\vspace{-3mm}
\label{tab:codec}
\begingroup
\setlength{\tabcolsep}{3pt}
\begin{tabular}{l l c c}
\specialrule{1.5pt}{1pt}{1pt}
Codec & Method &mIoU(\%) & GFLOPs \\
\toprule
\multirow{2}{*}{HEVC@3Mbps}
& PSPNet18(1.0x) & 69.43 & 309.02 \\
& AR$^{0.5}$-PSP18   & \textbf{70.48} & \textbf{101.98} \\
\hline
\multirow{2}{*}{HEVC@1Mbps}
& PSPNet18(1.0x)   & 65.76 & 309.02 \\
& AR$^{0.5}$-PSP18   & \textbf{67.89} & \textbf{101.98} \\
\hline
\multirow{2}{*}{x265-\textit{medium}@3Mbps}
& PSPNet18 (1.0x)  & 68.19 & 309.02 \\
& AR$^{0.5}$-PSP18 & \textbf{69.53} & \textbf{101.98} \\
\hline
\multirow{2}{*}{x265-\textit{ultrafast}@3Mbps}
& PSPNet18 (1.0x)  & 67.69 & 309.02 \\
& AR$^{0.5}$-PSP18 & \textbf{68.78} & \textbf{101.98} \\

\specialrule{1.5pt}{1pt}{1pt}

\end{tabular}
\endgroup
\vspace{-1mm}
\end{table}

\noindent \textbf{Video Compression Configurations.} 
As shown in Table \ref{tab:codec}, we also train and evaluate our model with different realistic bit-rates (3Mbps and 1Mbps) and configurations for HEVC/H.265 encoders. We use x265-\textit{medium} and x265-\textit{ultrafast} to represent different presets for x265, which apply simplified motion search algorithms and larger macro-blocks. These configurations are widely used in traditional video encoders. The results show that AR$^{0.5}$-PSP18 consistently outperforms the 1.0x constant resolution counterpart using only 33\% GFLOPs under different configurations.

\subsection{Running Time}

We measure the running time of AR-Seg with PSPNet18 on both CamVid and Cityscapes datasets, and the results are reported in Table \ref{tab:time}. Our AR-Seg models run 2x-3x times faster than the constant resolution counterparts. All tests are executed on a single GeForce RTX 3090 GPU.

\begin{table}[h]
\small
\centering
\caption{Running time of AR-PSP18 on 720x960 CamVid and 1024x2048 Cityscapes datasets.}
\vspace{-3mm}
\label{tab:time}
\begingroup
\setlength{\tabcolsep}{3.5pt}
\begin{tabular}{l | c | c | c}
\specialrule{1.5pt}{1pt}{1pt}
 Dataset & 1.0x baseline & AR$^{0.5}$ & AR$^{0.3}$ \\
\toprule
CamVid & 31.2 ms (32fps) & 14.7 ms (68fps) & 9.0 ms (111fps)\\
\hline
Cityscapes & 95.4 ms (10fps) & 30.7 ms (33fps) & 19.9 ms (50fps)\\

\specialrule{1.5pt}{1pt}{1pt}

\end{tabular}
\endgroup

\end{table}


\section{Conclusion}

In this paper, we propose AR-Seg, an altering resolution framework for compressed video semantic segmentation, which innovatively improves the efficiency of video segmentation from the perspective of input resolution. By jointly considering the design of architecture and training strategy, our proposed CReFF module and FST strategy effectively prevent the accuracy loss caused by downsampling. Results evaluated on two widely used datasets show that AR-Seg can achieve competitive segmentation accuracy with a reduction of up to 67\% computational cost. Our current study only uses two alternating resolutions (i.e., 
HR and LR). Future work that applies more complicated scheduling of multi-resolutions and keyframe gaps will be considered to further improve the VSS performance.

\vspace{1mm}
\noindent \textbf{Aknowledgements:}
This work was partially supported by Beijing Natural Science Foundation (L222008), Tsinghua University Initiative Scientific Research Program, the Natural Science Foundation of China (61725204), BNRist, and MOE-Key Laboratory of Pervasive Computing.

{\small
\bibliographystyle{ieee_fullname}
\bibliography{egbib}
}

\end{document}